\documentclass[10pt,twocolumn,letterpaper]{article}

\usepackage{iccv}
\usepackage{times}
\usepackage{booktabs}
\usepackage{epsfig}
\usepackage{graphicx}
\usepackage{amsmath}
\usepackage{amssymb}
\usepackage{bm}
\usepackage{subfigure}
\usepackage{multirow}
\usepackage{array}
\usepackage{enumerate}
\usepackage{mathrsfs}
\usepackage{amsmath}
\usepackage{amssymb}

\usepackage[breaklinks=true,bookmarks=false]{hyperref}

\iccvfinalcopy 

\def\iccvPaperID{8812} 

\ificcvfinal\fi

\begin{document}

\title{Instance Similarity Learning for Unsupervised Feature Representation}

\author{Ziwei Wang\textsuperscript{1,2,3},
	Yunsong Wang\textsuperscript{1},
	Ziyi Wu\textsuperscript{1},
	Jiwen Lu\textsuperscript{1,2,3}\thanks{Corresponding author},
	Jie Zhou\textsuperscript{1,2,3}\\	
	\textsuperscript{1} Department of Automation, Tsinghua University, China\\
	\textsuperscript{2} State Key Lab of Intelligent Technologies and Systems, China\\
	\textsuperscript{3} Beijing National Research Center for Information Science and Technology, China\\
	{\tt \small \{wang-zw18, wangys16\}@mails.tsinghua.edu.cn; dazitu616@gmail.com;}\\
	{\tt \small\{lujiwen,jzhou\}@tsinghua.edu.cn}
}

\maketitle
\ificcvfinal\thispagestyle{empty}\fi

\begin{abstract}
	In this paper, we propose an instance similarity learning (ISL) method for unsupervised feature representation. Conventional methods assign close instance pairs in the feature space with high similarity, which usually leads to wrong pairwise relationship for large neighborhoods because the Euclidean distance fails to depict the true semantic similarity on the feature manifold. On the contrary, our method mines the feature manifold in an unsupervised manner, through which the semantic similarity among instances is learned in order to obtain discriminative representations. Specifically, we employ the Generative Adversarial Networks (GAN) to mine the underlying feature manifold, where the generated features are applied as the proxies to progressively explore the feature manifold so that the semantic similarity among instances is acquired as reliable pseudo supervision.  Extensive experiments on image classification demonstrate the superiority of our method compared with the state-of-the-art methods. The code is available at \href{https://github.com/ZiweiWangTHU/ISL.git}{https://github.com/ZiweiWangTHU/ISL.git.}
	
\end{abstract}



\section{Introduction}
Deep neural networks have achieved the state-of-the-art performance in various vision applications such as face recognition \cite{deng2019arcface,wen2016discriminative,liu2017sphereface}, object detection \cite{ren2015faster,liu2016ssd,law2018cornernet}, image retrieval \cite{gordo2016deep,revaud2019learning,liu2016deep} and many others. However, most successful deep neural networks are trained with strong supervision, which requires a large amount of labeled data with expensive annotation cost and strictly limits the deployment of deep models. Hence, it is desirable to train deep neural networks with only the unlabeled data while achieving comparable performance with supervised learning.

\begin{figure}[t]
	\centering
	\includegraphics[height=5cm, width=8.3cm]{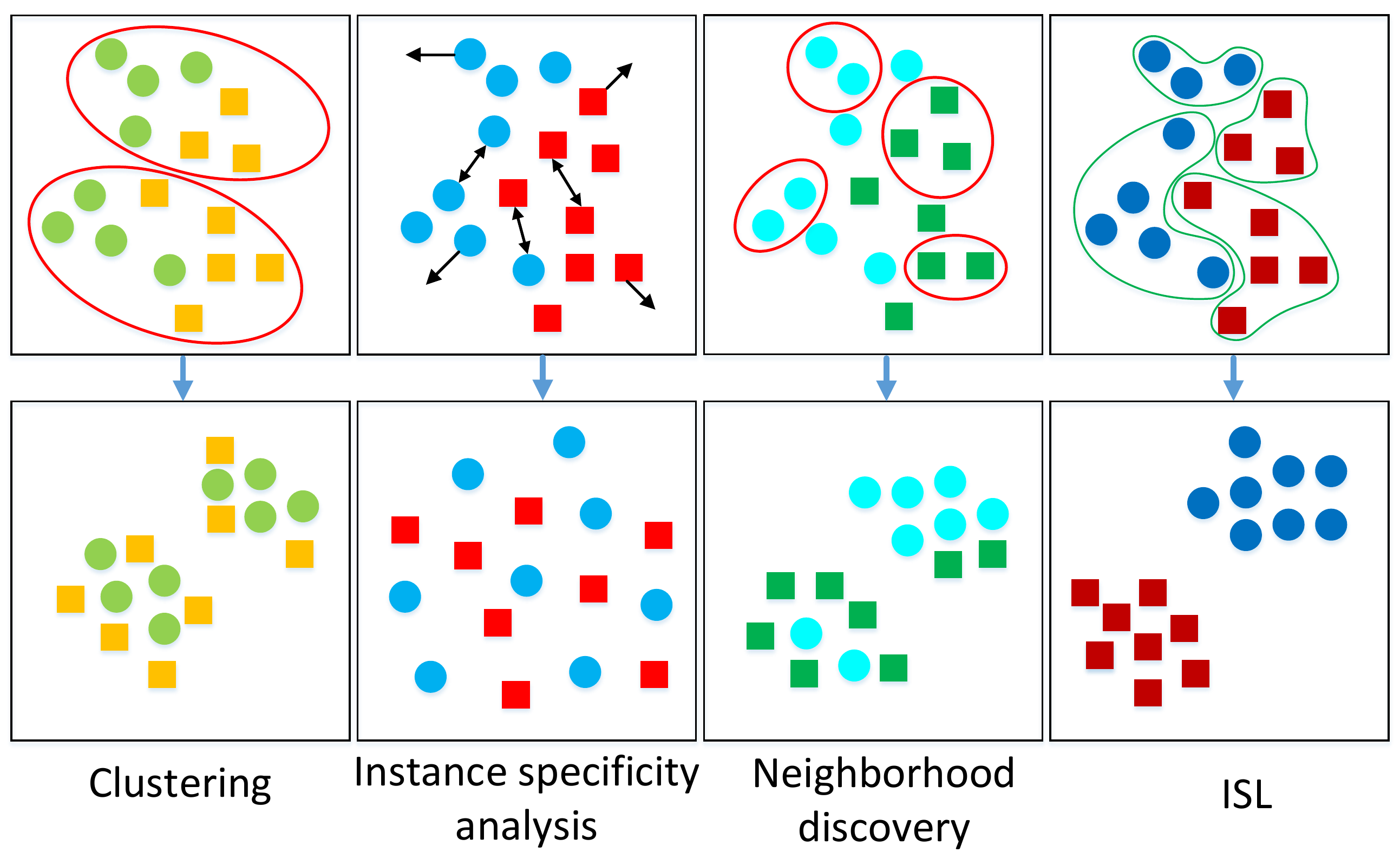}
	\vspace{-0.2cm}
	\caption{The difference among clustering methods, instance specificity analysis methods, neighborhood discovery methods and our method. The clustering methods are error-prone because of the complicated inter-class boundaries, and the instance specificity analysis methods are weakly discriminative due to the ambiguous supervision that treats each sample as an independent class. Meanwhile, the neighborhood discovery methods regard the instances close to the anchor as similar samples and fails to depict the true semantic similarity in large neighborhoods on the feature manifold. On the contrary, we mine the feature manifold and learn the instance-to-instance relationship with reliable semantic similarity, so that informative features can be obtained.}
	\label{fig:example}
	\vspace{-0.5cm}
\end{figure}

To enable deep neural networks to learn from the unlabeled data, unsupervised learning methods have been widely studied recently. The clustering methods \cite{ji2018invariant,xie2016unsupervised,caron2018deep} shown in the first column of Figure \ref{fig:example} provide pseudo labels to train the networks according to the cluster indexes, which are error-prone due to the complex inter-class boundaries. The instance specificity analysis methods  \cite{wu2018unsupervised,bojanowski2017unsupervised,oord2018representation,he2019momentum,hjelm2018learning} depicted in the second column of Figure \ref{fig:example} regard every single sample as an independent class to avoid clustering. However, the offered supervision is ambiguous and results in weak class discrimination. Meanwhile, designing pretext tasks with self-supervised learning \cite{doersch2015unsupervised,zhang2017split,wang2015unsupervised} shares the same limitations of instance specificity analysis methods due to the discrepancy between the auxiliary supervision and the target task. In order to mitigates the disadvantages of clustering and instance specificity analysis, neighborhood discovery methods \cite{huang2019unsupervised,zhuang2019local,huang2020unsupervised} have been proposed, which explore the local neighbors progressively with class consistency maximization by mining instance-to-instance correlation. They simply assign high similarity to pairs that have short Euclidean distance in the feature space. While the representations lie in the implicit feature manifold that is continuous in the Euclidean space, the Euclidean distance only reveals the true semantic similarity in extremely small neighborhoods and fails to provide the informative pseudo supervision for large neighborhoods due to the inconsistency with the distance measured on the feature manifold. As a result, the feature discriminality is still limited as shown in the third column of Figure \ref{fig:example}.

\begin{figure*}[t]
	\centering
	\includegraphics[height=5.2cm, width=15cm]{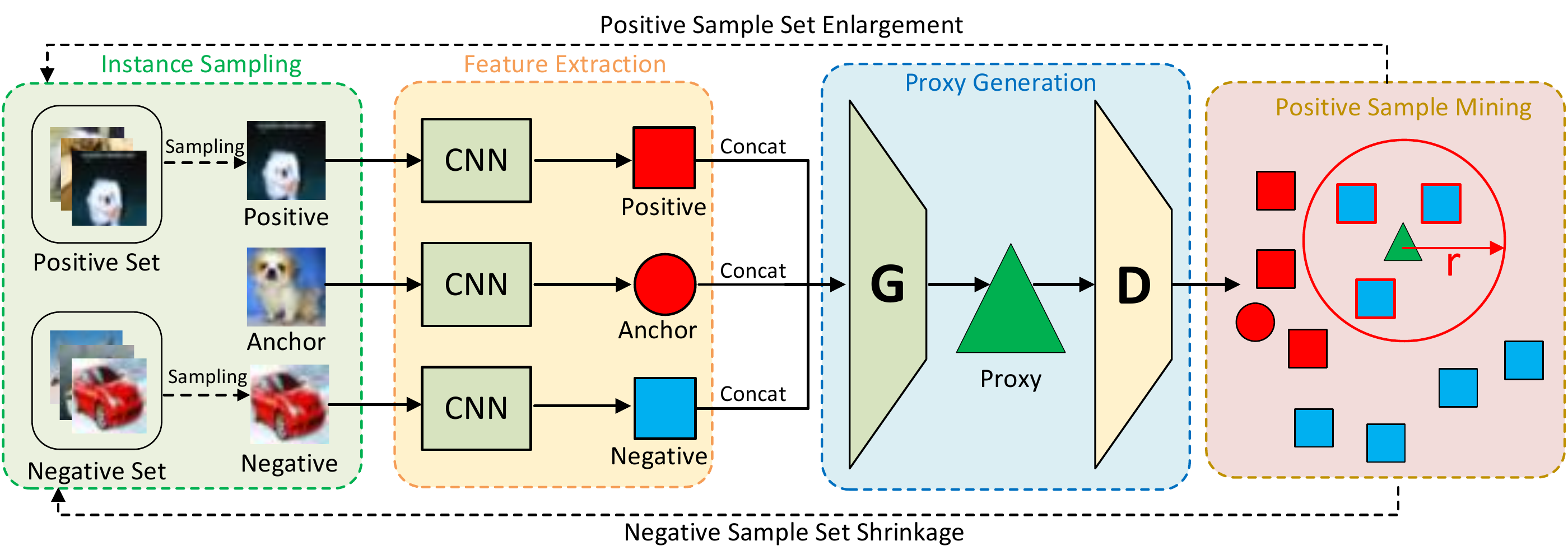}
	\vspace{-0.1cm}
	\caption{The pipeline of the instance similarity learning. For a given anchor, we first sample triplets from the mined positive set and the negative set, and then obtain the features via the convolutional neural networks. After concatenating the features of the anchor, the positive and the negative samples, we generate the proxy for feature manifold mining by the generator. The instances in the neighborhood of the proxy are removed from the negative set and added to the positive set if the proxy is semantically similar to the anchor, where the semantics similarity is predicted by the discriminator.}
	\label{fig:pipeline}
	\vspace{-0.5cm}
\end{figure*}

In this paper, we present an ISL method to learn the semantic similarity among instances for unsupervised feature representation. Unlike the conventional methods that assign high similarity to close pairs according to the Euclidean distance in the feature space, our method mines the feature manifold in an unsupervised manner and learns the semantic similarity among different samples, so that the reliable instance-to-instance relationship in large neighborhoods is applied to supervise the representation learning models as demonstrated in the last column of Figure \ref{fig:example}. More specifically, we employ the Generative Adversarial Network (GAN) \cite{goodfellow2014generative} to mine the underlying feature manifold, and Figure \ref{fig:pipeline} depicts the overall pipeline of the proposed method. The generator yields the proxy feature that mines positives for each anchor instance based on the sampled triplet, and the discriminator predicts the confidence score that the generated proxy is semantically similar with the mined pseudo positive samples. Since the Euclidean distance reveals the sample similarity in small neighborhoods, the instances near the proxy feature with high confidence score are added to the positive sample set of the given anchor. In order to explore richer instance-wise relation and exploit the semantics of the mined positive sample set simultaneously, the generated proxy is enforced to be similar with negative instances and the mined pseudo positive samples during the training process of GANs. With the reliable pseudo supervision, we employ the contrastive loss with hard positive enhancement to learn discriminative features. Extensive experiments on CIFAR-10 \cite{krizhevsky2009learning}, CIFAR-100 \cite{krizhevsky2009learning}, SVHN \cite{netzer2011reading} and ImageNet \cite{deng2009imagenet} datasets for image classification demonstrate that the proposed ISL outperforms most of the existing unsupervised learning methods. Moreover, our ISL can be integrated with state-of-the-art unsupervised features to further enhance the performance.

\section{Related Work}
Unsupervised learning has aroused extensive interests because it enables models to be trained by vast unlabeled data and saves expensive annotation cost. Existing methods can be divided into five categories: clustering, instance specificity analysis, neighborhood discovery, self-supervised learning and generative models.

\textbf{Clustering:} Clustering methods \cite{caron2018deep,xie2016unsupervised,ji2018invariant,yang2017towards} employ cluster indexes as pseudo labels to train the end-to-end unsupervised learning model. Caron \etal \cite{caron2018deep} jointly learned the network parameters and the cluster assignment of features, where k-means was applied for iterative data grouping. Furthermore, Yang \etal \cite{yang2017towards} applied stacked autoencoders \cite{vincent2010stacked} to provide stronger supervision by minimizing the image reconstruction loss despite of the cluster assignment. However, the clustering methods are error-prone as they fail to represent the highly complex class boundaries.

\textbf{Instance specificity analysis:} Instance specificity analysis \cite{wu2018unsupervised,bojanowski2017unsupervised,oord2018representation,bachman2019learning,he2019momentum,hjelm2018learning,chen2020simple,zhang2019aet, henaff2019data, ye2019unsupervised} methods consider every single instance as an independent class, and only take the sample and its transformed instance as positive pairs with the assumption that the instance semantic similarity is automatically discovered with the instance-wise supervision. Wu \etal \cite{wu2018unsupervised} proposed the noise-contrastive estimation (NCE) to approximate the full softmax distribution in order to reduce the complexity of the instance-wise classifier, and utilized a memory bank to store the instance feature. He \etal \cite{he2019momentum} built dynamic dictionary on-the-fly that facilitated largescale contrastive learning. Chen \etal \cite{chen2020simple} composed various data augmentation techniques with an extra non-linear transformation to learn discriminative unsupervised features. However, the learned class boundaries are ambiguous in instance specificity analysis methods as they may push away samples with the same class label and increase the intra-class variance.

\textbf{Neighborhood discovery: }The neighborhood discovery methods \cite{huang2019unsupervised, huang2020unsupervised,zhuang2019local} mitigate the drawbacks of the above two kinds of methods by progressively mining instance-to-instance correlation with class consistency maximization. Huang \etal \cite{huang2019unsupervised} iteratively enlarged the neighborhood for each instance by comparing its cosine similarity with different samples in the curriculum learning setting, and treated all neighbors as positive instances. Zhuang \etal \cite{zhuang2019local} presented a metric for local aggregation, where similar samples were encouraged to move together and vice versa. Nevertheless, existing neighborhood discovery methods simply assign the similarity based on the Euclidean distance of their features to train the representation learning model, which fails to demonstrate the semantic similarity on the underlying feature manifold for large neighborhoods. 

\textbf{Self-supervised learning: }self-supervised learning methods \cite{doersch2015unsupervised,zhang2017split,wang2015unsupervised,noroozi2016unsupervised,pathak2016context,kim2019self, misra2020self, feng2019self} usually design pretext tasks to provide the hand-crafted auxiliary supervision with human priors, where the assumption is that the semantics learned via the auxiliary supervision can be transferred to the downstream tasks such as image classification and object detection. Doersch \etal \cite{doersch2015unsupervised} and Noroozi \etal \cite{noroozi2016unsupervised} sampled patches on a image and designed the jigsaw puzzles, where the networks were designed to predict the relative position of two patches. Pathak \etal \cite{pathak2016context} used the context-based pixel prediction as the pretext task, and the masked contents in an image should be generated by the context encoders with reconstruction and adversarial loss. However, the self-supervised learning methods share the same limitations with the instance specificity analysis methods in unsupervised learning due to the large discrepancy between the pretext tasks and the downstream applications.

\textbf{Generative Models: }Generative models \cite{vincent2010stacked,lee2009convolutional,kingma2013auto,radford2015unsupervised,hinton2006fast,goodfellow2014generative, donahue2019large} including RBM \cite{hinton2006fast}, AutoEncoders \cite{kingma2013auto} and GAN \cite{goodfellow2014generative} have been widely studied recently since it is able to learn the data distribution by reconstructing the input samples without supervision. Radford \etal \cite{radford2015unsupervised} and Donahue \etal \cite{donahue2016adversarial} applied the GANs to extract representations that generated samples semantically similar to the input. Learning representations directly with generative models leads to weak class discriminality due to the difference between the reconstruction and downstream tasks.

\section{Approach}
In this section, we first introduce feature manifold mining via GANs, and then present the instance semantic similarity learning on the mined feature manifold. Finally, we propose effective training objective with the learned semantic similarity to obtain discriminative representations.

\subsection{Feature Manifold Mining}

Let $X=\{\bm{x}_1, \bm{x}_2,...,\bm{x}_N\}$ and $F=\{\bm{f}_1, \bm{f}_2,...,\bm{f}_N\}$ be the input images and their features respectively, where $N$ is the number of instances. $S\in\{0,1\}^{N\times N}$ is the similarity matrix, where the element in the $i_{th}$ row and $j_{th}$ column $s_{ij}$ equals to one if $\bm{x}_i$ and $\bm{x}_j$ are semantically similar (positive) and zero otherwise (negative). Conventional unsupervised methods treat pairs with short Euclidean distance in the feature space as similar ones. However, the Euclidean distance only reveals the similarity in extremely small neighborhoods and usually fails to depict the true semantic similarity in large neighborhoods due to the mismatch between the geodesic distance on the feature manifold and the Euclidean distance. As a result, samples with dissimilar semantics are regarded as similar pairs for pseudo supervision to train the representation model and vice versa, which leads to uninformative features in unsupervised learning. Because the implicit feature manifold changes during the training process of the feature extraction model, we employ GANs to dynamically mine the feature manifold according to the feature distribution.

In order to evaluate feature distribution, we sample the triplets $\{\bm{f}_i^a, \bm{f}_{i}^p, \bm{f}_{i}^n\}$ in the feature space according to the similarity matrix $S$, where $\bm{f}_i^a$, $\bm{f}_{i}^p$ and $\bm{f}_{i}^n$ are the features of the anchor, the positive sample and the negative sample in the $i_{th}$ triplet respectively. For initialization, $S$ is set to be the identity matrix at the beginning of training, which means that all instances are only semantically similar with themselves. The proxy generator $G$ generates the proxy feature $\bm{f}_{i}^g$ for the $i_{th}$ triplet that is used to explore the feature manifold by mining positives for the given anchor and modify the similarity matrix dynamically. Aiming to explore richer instance-wise relation and exploit the semantics of the mined positive sample set simultaneously, we expect the proxy feature $\bm{f}_{i}^g$ to have the two following properties:

\begin{enumerate}[(1)]
	\item The proxy feature should be semantically similar with negative samples in the triplet. At the beginning of training process, the positive sample in the triplet is identical with the anchor, where the rich instance-wise relation is not explored for discriminative representation learning. In order to enlarge the positive sample set for more informative supervision, enforcing the proxy to be semantically similar to negatives enables active feature manifold exploration.
	\item The proxy feature should also be semantically similar to positive samples with the goal of exploiting the semantics from mined positive sets, so that the feature manifold is learned with high precision.
\end{enumerate}

We employ a discriminator $D$ to measure the semantic similarity between the proxy and the positives or negatives. $D$ should accurately classify the real triplet $\mathcal{T}_r=\{\bm{f}_i, \bm{f}_{i}^p, \bm{f}_{i}^n\}$ sampled from the mined sets and synthetic triplet $\mathcal{T}_{s}^n=\{\bm{f}_i, \bm{f}_{i}^p, \bm{f}_{i}^g\}$ with the generated proxy as the negative. Meanwhile, the real triplet should also be distinguished by $D$ from the synthetic triplet $\mathcal{T}_{s}^p=\{\bm{f}_i, \bm{f}_{i}^g, \bm{f}_{i}^n\}$ with the generated proxy as the positive. Following the adversarial loss in \cite{goodfellow2014generative}, we design the following objective to train the generator and discriminator, and obtain the proxy feature similar to both positive and negative samples:
\begin{align}\label{gan_loss}
	\min\limits_{G}\max\limits_{D} \mathcal{L}_{gan}=&\log D(\mathcal{T}_r)+\log(1-D(\mathcal{T}_{s}^p))+\\\notag
	&\alpha\log(1-D(\mathcal{T}_{s}^n))
\end{align}where $\bm{f}_{i}^g$ in $\mathcal{T}_{s}^n$ and $\mathcal{T}_{s}^p$ is generated by $G$ based on the real triplet $\mathcal{T}_r$ and is denoted as $\bm{f}_{i}^g=G(\mathcal{T}_r)$. $D(\mathcal{T})$ represents the confidence score that the input triplet $\mathcal{T}$ is real, which is predicted by the discriminator. $\alpha$ is a hyperparameter that balances the hardness of the generated proxy feature to be recognized as positive samples. When $\alpha$ increases, the generated proxy $\bm{f}_{i}^g$ is forced to be more similar to the negative sample and is harder to be recognized as the positive instance, which means the proxy explores the feature manifold more aggressively. When finishing the training of GANs, the generator $G$ learns the underlying feature manifold and is able to generate the reliable proxy to enlarge the positive sample set for the given anchor.

\subsection{Instance Similarity Learning}
In this section, we first briefly introduce the hand-crafted instance similarity assignment in conventional methods that utilize the Euclidean distance among features to measure the similarity, and then detail the instance similarity learning with the mined feature manifold in our method. In conventional methods \cite{huang2019unsupervised,huang2020unsupervised,zhuang2019local}, the neighborhood $\mathcal{N}(\bm{x})$ is identified by k-nearest neighbors for a given anchor $\bm{x}$ in the following form:
\begin{align*}
	\mathcal{N}(\bm{x})=\{\bm{x}_i|d(\bm{x}_i,\bm{x}) \rm ~is~ ranked~ the~ bottom~\emph{k}~ for~ all~ \emph{i}\}
\end{align*}where $d(\bm{x},\bm{y})$ means the distance between two feature vectors $\bm{x}$ and $\bm{y}$, and the Euclidean distance is usually applied. $k$ is a hyperparameter that decides the size of the neighborhood, and instances in the neighborhood are all treated as similar samples. Since the Euclidean distance can only reveal the true semantic similarity in extremely small neighborhoods and fails to provide informative pseudo supervision in large neighborhoods, $k$ is usually limited to be very small and the class discrimination is weak due to the constrained size of the positive sets.

Our method employs the generated proxy $\bm{f}_{i}^g$ to mine the semantically similar instances with the anchor feature $\bm{f}_i$ in order to enlarge the positive sample set $\mathcal{P}_i$, where $\mathcal{P}_i$ is initialized with the anchor itself. Since the generator $G$ learns the underlying feature manifold according to the feature distribution, the generated proxy $\bm{f}_{i}^g$ is utilized to mine semantically similar instances for the given anchor and move the semantically similar samples from the negative set to enlarge the positive one. Because the confidence score of the synthetic triplet $D(\mathcal{T}_{s}^p)$ evaluates the sematic similarity between the generated feature and the mined positives, it represents the reliability of the proxy for positive sample set enlargement. When the confidence score $D(\mathcal{T}_{s}^p)$ of the generated proxy is high, the proxy mines the reliable region in which the instances are removed from the negative sample set and added to the positive one. We employ the following strategy to enlarge the positive sample set $\mathcal{P}_i$ for a given anchor $\bm{f}_i$ with the instance $\bm{f}_j$:
\begin{align}\label{enlarge}
	\bm{f}_j=\{\bm{f}_j\big| ||\bm{f}_{i}^{g}-\bm{f}_{j}||_F<r,~D(\mathcal{T}_{s}^p)>h\}
\end{align}where $||\cdot||_F$ means the Frobenius norm and $r$ is a hyperparameter to control the size of the region for positive sample set enlargement. $h$ is the threshold to trigger positive sample addition. Since the feature manifold is continuous in the feature space, the Euclidean distance can reveal the semantic similarity in extremely small neighborhoods. As a result, instances in the small hyperspherical neighborhoods of the proxy can be treated as semantically similar samples with the proxy feature, which share consistent semantics with the anchor for positive sample set enlargement. 

Because the generated proxy $\bm{f}_{i}^g$ is influenced by the sampled real triplet $\mathcal{T}_{r}$ input to the generator $G$, we sample the real triplets of the given anchor for multiple times to gain more information about the distribution of the positives and negatives. We denote the optimal proxy as $\bm{f}_{i}^{g*}$ with the definition in the following:
\begin{align}\label{optimal}
	\bm{f}_{i}^{g*}=\arg \max \limits_{\bm{f}_{i}^g} D(\mathcal{T}_{s}^p)
\end{align}We utilize the optimal proxy among all generated proxy features to enlarge the positive sample set via (\ref{enlarge}). The pseudo supervision provided by instance similarity learning is informative as it sets the instances with short geodesic distance on the mined feature manifold to be positive, and maximizing the similarity between their features can significantly enhance the feature informativeness on downstream tasks such as image classification and object detection.

\subsection{Learning Representations with the Mined Instance Similarity}
The learned instance similarity can provide effective supervision for unsupervised feature representation, where the semantic similar pairs should be constrained to be close in the feature space and vice versa. Following non-parametric loss in \cite{wu2018unsupervised}, we illustrate the similarity by the probability distributions $p_{ij}$ that two samples $\bm{x}_i$ and $\bm{x}_j$ come from the same class: 
\begin{align}\label{probability}
	p_{ij}=\frac{\exp(\bm{f}_i^T\bm{f}_j/\tau)}{\sum_{k=1}^{N}{\exp(\bm{f}_i^T\bm{f}_k/\tau)}}
\end{align}where $\tau$ is the hyperparameter for the temperature that controls the concentration of the distribution \cite{hinton2015distilling}. Since we argue that all semantically similar instances in the positive sample set $\mathcal{P}_i$ for a given anchor share the same class label, we propose the following objective that maximizes the log-likelihood of the probability that all instances in the positive sample set come from the same class:
\begin{align}\label{objective} 
	\mathcal{L}_1=-\sum_{i=1}^{N}\log(\sum_{\bm{f}_k\in\mathcal{P}_i}p_{ik})
\end{align}The objective is to encourage the label consistency between the anchor and all of its positive samples, so that the more informative pseudo supervision for representation learning is provided. 

As demonstrated in \cite{huang2020unsupervised}, the less semantically similar instances in the positive sample set can be overwhelmed during training because of the small quantity. However, the hard positives provide large gradients and contribute significantly to the training process  \cite{zhao2018adversarial,zheng2019hardness,harwood2017smart}. As a result, we apply the hard positive enhancement (HPE) strategy demonstrated in \cite{huang2020unsupervised} to further enhance the performance. We define the positive sample $\bm{f}_j$ with smallest $p_{ij}$ w.r.t. the anchor $\bm{f}_i$ as the hard positive. For initialized positive sample set, the feature of a randomly transformed variant of the anchor image $\bm{x}_i$ is regarded as hard positive. Denoting the hard positive of the anchor $\bm{f}_i$ as $\bm{f}_i^{hard}$, we employ the following loss to integrate the hard positive enhancement strategy with our method:
\begin{align}\label{hard_positive}
	\mathcal{L}_2=\sum_{i=1}^{N}\sum_{k=1}^{N}p_{ik}\log\frac{p_{ik}}{p_{ik}^{hard}}
\end{align}where $N$ is the number of samples in the dataset, and $p_{ik}^{hard}$ demonstrates the probability that the instance $\bm{f}_k$ and the hard positive $\bm{f}_i^{hard}$ of the anchor $\bm{f}_i$ comes from the same class. The loss for hard positive enhancement significantly magnifies the influence of hard positives during training, which leads to discriminative boundaries among classes in the feature space. The overall loss for our ISL is written as follows:
\begin{align}\label{overall_loss}
	\mathcal{L}=\mathcal{L}_1+\lambda\mathcal{L}_2
\end{align}where $\lambda$ is a hyperparameter that balances the importance of two loss terms. For fair comparison with the state-of-the-art methods, we conducted experiments in the settings with and without the hard positive enhancement strategy. Following \cite{wu2018unsupervised}, we maintain an offline memory bank to avoid intractable loss computations for all the instances by storing feature vectors in the memory. We initialize the memory bank with random vectors and update the memory features $\hat{\bm{f}_i}$ by mixing the memory features and the learned up-to-date features $\bm{f}_i$:
\begin{align}\label{memory_bank}
	\hat{\bm{f}_i}=\eta\bm{f}_i+(1-\eta)\hat{\bm{f}_i}
\end{align}where $\eta\in[0,1]$ is a hyperparameter that illustrates the importance of up-to-date features during the process of memory update.

\section{Experiments}
In this section, we first describe the datasets and our implementation details briefly. Then we demonstrate our intuitive logic by toy examples, and conducted the ablation study to investigate the impact of different components in the presented instance similarity learning. Finally, we compare our ISL with the state-of-the-art unsupervised feature learning methods on image classification. The implementation details and the results on other tasks such as object detection and transfer learning are shown in the supplementary material.

\subsection{Datasets and Implementation Details}
We first detail the datasets that we carried out experiments on: The CIFAR-10 dataset consists of $60,000$ images from $10$ classes with $50,000$ images for training and $10,000$ for evaluation. The CIFAR-100 dataset has the same data split with CIFAR-10, and the only difference is the images consist of 100 classes with 600 images for each. The Street View House Numbers (SVHN) dataset contains $10$ classes of digit images with $73,257$ of them for training and $26,032$ of them for evaluation. The ImageNet dataset consists of about $1.2$ million and $50k$ images from $1,000$ classes for training and validation respectively. 

We employed the top-1 accuracy to evaluate ISL on image classification. Following the experiment settings in \cite{wu2018unsupervised}, we tested two classifiers including Linear Classifier (LC) and Weighted kNN to evaluate the features extracted in different layers. We applied a fully-connected layer as the LC, which was trained by the cross-entropy loss. The weighted kNN classifier infers the class label for the feature $\bm{f}$ by the votes of the top-$k$ neighbors. For each neighbor $\bm{f}_i$, the weight is assigned to be $\exp(\bm{f}_i^T\bm{f}/\tau)$. We set $k=200$ and $\tau=0.07$ in our experiments. We trained our ISL with the architectures of the AlexNet \cite{krizhevsky2012imagenet}, ResNet18 and ResNet50 \cite{he2016deep}. 

We iteratively trained GANs that mined the feature manifold, learned the semantic similarity among instances and optimized the backbone that extracted unsupervised features of images with $4$ rounds in total. In the training of GANs, we sampled five triplets for a given anchor in order to decrease the discrepancy between the sampled feature distribution and the real feature distribution so that the feature manifold could be mined precisely. We leveraged three fully-connected layers as the generator and another three-layer fully-connected networks as the discriminator. In each round, we trained GANs until the loss of the generator converged. The hyperparameter $\alpha$ was set to $1$. We used the Adam optimizer \cite{kingma2014adam} with fixed learning rate $1e$-$4$ to train both the generator and discriminator.

In the training of the backbone networks, the number of training epochs in each round was $200$, $200$, $100$ and $100$ for experiments on CIFAR-10, CIFAR-100, SVHN and ImageNet respectively. Following \cite{wu2018unsupervised}, we adopted the SGD optimizer with momentum at $0.9$. The learning rate was initially set to $0.03$ and decayed twice by multiplying $0.1$ at the $75\%$ and $90\%$ epoch of the total epochs. We used a batchsize of $256$ for ImageNet and $128$ for others. The feature was normalized and the length was fixed to 128 in most experiments. The hyperparameter $\eta$, $\tau$ and $\lambda$ were set as $0.5$, $0.07$ and $0.5$ respectively.

For instance similarity learning that enlarges the positive sample set, we sampled five triplets for a given anchor to generate the optimal proxy feature with enhanced reliability. The hyperparameters $h$ and $r$ were set to $0.5$ and $1$.

\begin{figure}[t]
	\centering
	\begin{center}
		\includegraphics[height=3.5cm, width=8.3cm]{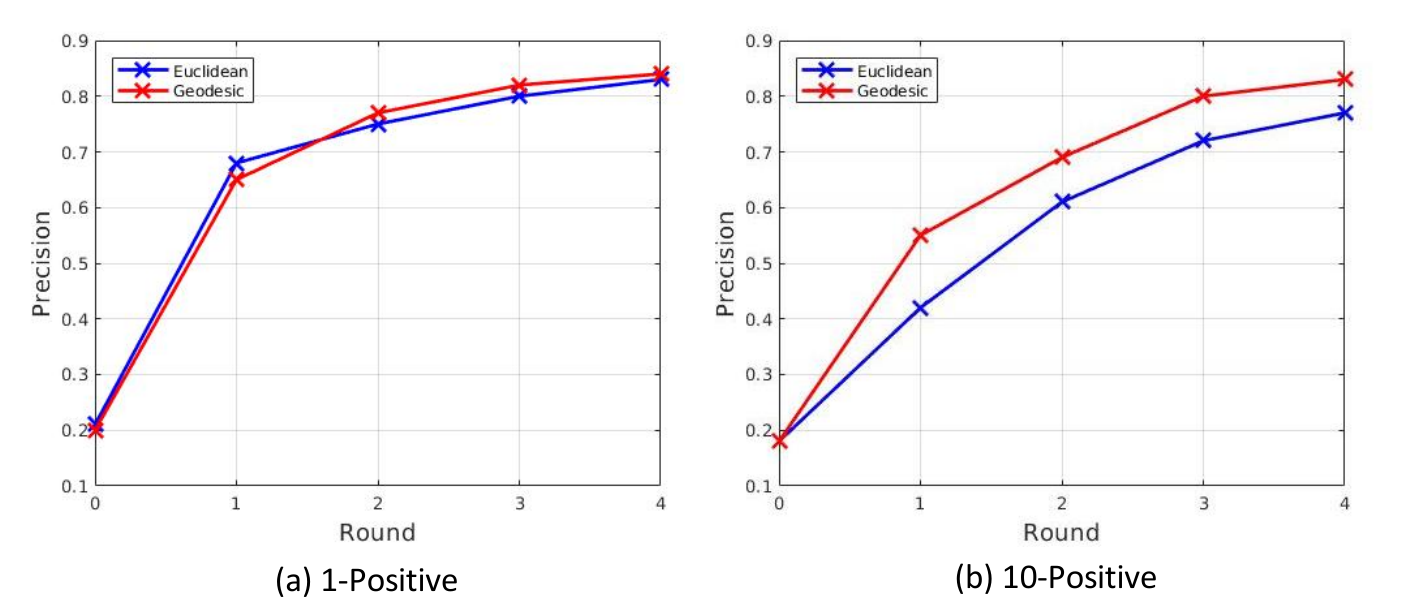}
	\end{center}
	\vspace{-0.3cm}
	\caption{The average precision of mined positive samples w.r.t. different rounds during training for positive sample set size of $1$ and $10$. }
	\label{toy}
	\vspace{-0.6cm}
\end{figure}

\subsection{Performance Analysis}
In this section, we first demonstrate the intuitive logic of our instance similarity learning by toy examples, and show the influence of different components in the proposed techniques by ablation study.

\vspace{-0.3cm}
\subsubsection{Toy Examples}
\vspace{-0.1cm}
While Euclidean distance fails to reveal the true semantic similarity for samples in large neighborhoods, the thought of the presented ISL is learning the instance similarity in the feature manifold to provide informative pseudo supervision for unsupervised feature representation. We conducted simple experiments on CIFAR-10 with AlexNet to show our thoughts with intuition.

We show the average precision of the positive sample sets across all anchors, where the precision is defined as the ratio of mined pseudo positives from the anchor class. Figure \ref{toy} demonstrates the precision of mined pseudo positives across anchors w.r.t. different epochs during training, where the positive sample set size was $1$ and $10$. The geodesic distance applied in our ISL is compared with the Euclidean distance leveraged in conventional neighborhood discovery methods, and the latter chose the closest samples to be positive. Both distance measure achieves similar precision for the positive sample set in size of $1$, while geodesic distance significantly surpasses Euclidean distance for the positive sample set in size of $10$ since the former reveals the true semantic similarity in large neighborhoods.

\vspace{-0.3cm}
\subsubsection{Ablation Study}
\begin{figure}[t]
	\centering
	\begin{center}
		\includegraphics[height=2.45cm, width=8.3cm]{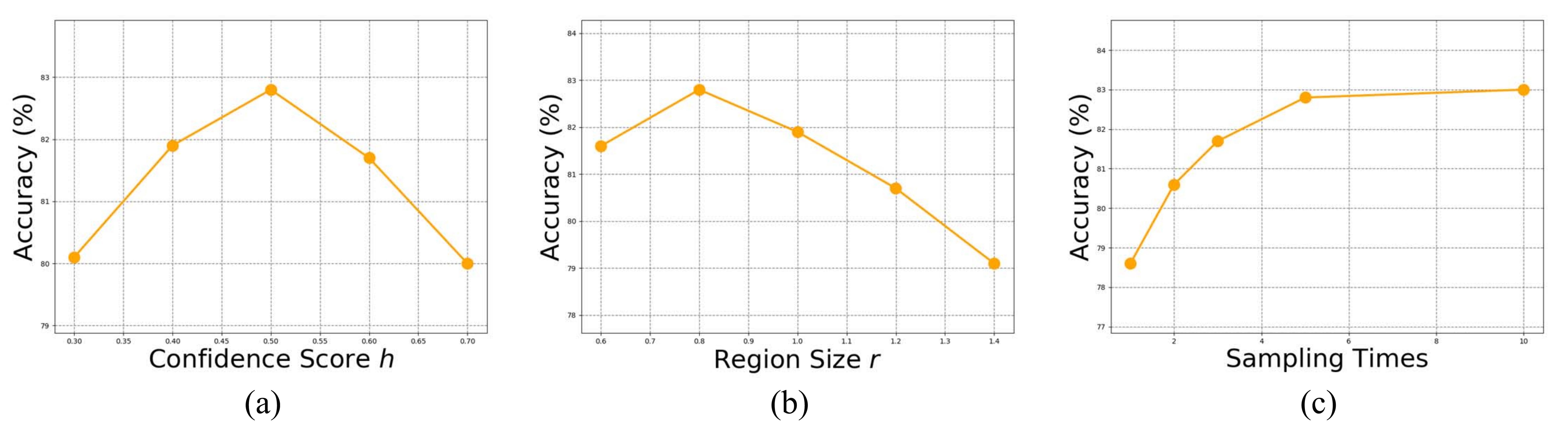}
	\end{center}
	\vspace{-0.3cm}
	\caption{Classification accuracy on the CIFAR-10 dataset w.r.t. (a) the confidence score threshold $h$ in instance similarity learning and (b) the region size $r$ and (c) the sampling times to generate each proxy for positive sample set enlargement.}
	\label{fig:ablation}
	\vspace{-0.6cm}
\end{figure}

\vspace{-0.1cm}
Leveraging the Euclidean distance among features as the supervision only reveals the semantic similarity in extremely small neighborhoods and fails to provide informative pseudo supervision for representation learning. On the contrary, our instance similarity learning illustrates geodesic distance on the mined feature manifold that demonstrates the reliable instance-to-instance relationship. In order to investigate the effectiveness of the proposed instance similarity learning and the impact of the critical hyperparameters, we conducted ablation study w.r.t. the confidence score threshold $h$ in instance similarity learning, the region size $r$ for positive sample set enlargement and the sampling times to generate each proxy for positive sample set enlargement. We adopted the AlexNet architecture as the backbone and trained our ISL on the CIFAR-10 dataset in the ablation study. The kNN classification accuracy is reported for evaluation, which is shown in Fig. \ref{fig:ablation}.

\textbf{Performance w.r.t. the confidence score threshold $h$: }In instance similarity learning, the generated proxy is applied to enlarge the positive sample set using the surrounding instances when the confidence score is larger than the threshold $h$. Increasing $h$ reduces the mined positives for the given anchor because the proxy is required to be more confident in positive sample set enlargement and vice versa. The impact of $h$ on the performance is illustrated in Fig. \ref{fig:ablation}(a), where medium threshold achieves the best performance. The low threshold is not able to guarantee the reliability of the generated proxy and the high threshold fails to provide sufficient proxies for positive sample set enlargement, where both of them degrade the accuracy.

\textbf{Performance w.r.t. the region size $r$: }In positive sample set enlargement, instances whose Euclidean distance from the proxy is less than $r$ are assigned to be the positives for the given anchor. Larger $r$ represents that more instances are added to the positive sample set for each generated proxy, and assumes that the Euclidean distance can better approximate the geodesic distance on the feature manifold in larger neighborhoods. Fig. \ref{fig:ablation}(b) demonstrates the performance versus different $r$, and medium $r$ enlarges the positive sample set with sufficient reliable instances. Large $r$ adds unreliable instances to the positive sample set because the Euclidean distance cannot reveal the true semantic similarity in large neighborhoods. On the contrary, insufficient instances are added to the positive sample set for small $r$, so that the samples with similar semantics are pushed away and the class boundaries of features become ambiguous.

\textbf{Performance w.r.t. the sampling times to generate the proxy: }The generator $G$ generates the proxy feature according to the anchor, the distribution of its positive samples and negative samples. In order to provide accurate information about the distribution, we sampled the triplets for multiple times so that the more reliable proxy could be generated. The performance w.r.t. different sampling times is illustrated in Fig. \ref{fig:ablation}(c), where the classification accuracy increases when the triplets are sampled for more times. However, the improvements become very incremental when the sampling time is larger than five, while the computational cost during the training stage increases significantly. To balance the efficiency and the effectiveness, we sampled five triplets to generate reliable proxies in most experiments.

\begin{figure*}[t]
	\centering
	\begin{center}
		\includegraphics[height=3.8cm, width=17cm]{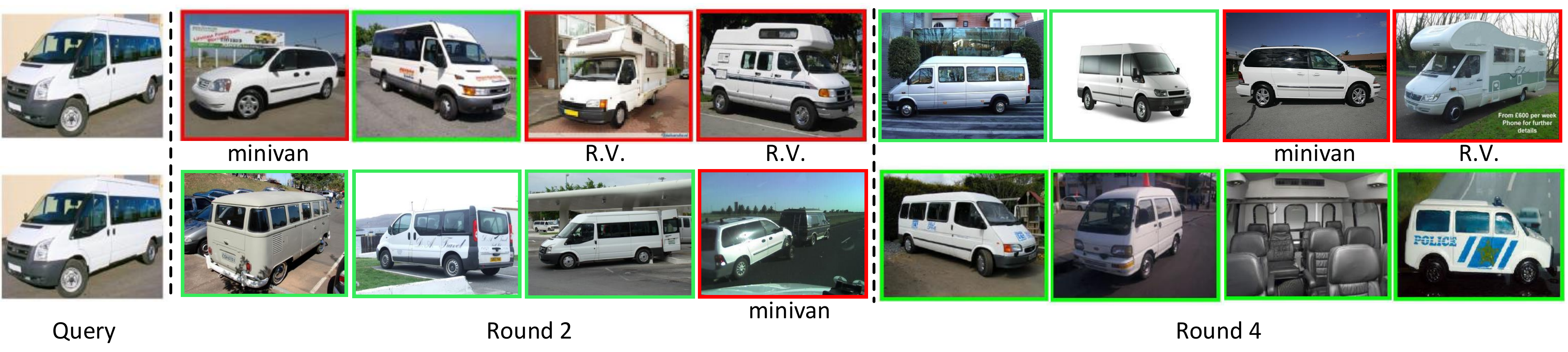}
	\end{center}
	\vspace{-0.5cm}
	\caption{An example of positive sample mining via LA (top row) and our ISL (bottom row) in different rounds during training. The query image is from the minibus class. The images with green boxes represent the positives mined correctly and those with red boxes mean the images from other classes. More examples are visualized in the supplementary material.}
	\label{fig:visualization}
	\vspace{-0.3cm}
\end{figure*}

\begin{table}[t]
	\footnotesize
	\caption{Classification accuracy (\%) on CIFAR-10, CIFAR-100 and SVHN, where the architecture of AlexNet, ResNet18 and ResNet50 were applied as the backbone networks. The results of two classification models are reported: the weighted kNN with the FC features and the linear classifier using the Conv5 features. ISL w/o HPE means our method without hard positive enhancement.}
	\label{small_dataset}
	\centering
	\vspace{0.2cm}
	\renewcommand\arraystretch{1.1}
	\begin{tabular}{c|c|ccc}
		\hline
		& Dataset & CIFAR10 & CIFAR100 & SVHN\\
		\hline
		Architecture & Classifier/Feat. & \multicolumn{3}{c}{Weighted $k$NN / FC}\\
		\hline
		\multirow{8}{*}{AlexNet} & Random & $34.5$ & $12.1$ & $56.8$\\
		\cline{2-5}
		& DeepCluster & $62.3$ & $22.7$ & $84.9$\\
		& RotNet & $72.5$ & $32.1$ & $77.5$\\
		& Instance & $60.3$ & $32.7$ & $79.8$\\
		& AND & $74.8$ & $41.5$ & $90.9$\\
		& ISL w/o HPE & $\textbf{81.1}$ & $\textbf{49.2}$ & $\textbf{91.0}$\\
		\cline{2-5}
		& PAD & $81.5$ & $48.7$ & $91.2$\\
		& ISL & $\textbf{82.8}$ & $\textbf{50.3}$ & $\textbf{91.8}$\\
		\hline
		\multirow{4}{*}{ResNet18} & Instance & $80.8$ & $40.1$ & $92.6$\\
		& AND & $86.3$ & $48.1$ & $93.1$\\
		& ISL w/o HPE & $\textbf{87.0}$ & $\textbf{52.1}$ & $\textbf{93.9}$\\
		\cline{2-5}
		& ISL & $\textbf{87.8}$ & $\textbf{54.7}$ & $\textbf{94.2}$\\
		\hline
		\multirow{4}{*}{ResNet50} & Instance & $81.8$ & $42.3$ & $92.9$\\
		& AND & $87.6$ & $49.0$ & $93.2$\\
		& ISL w/o HPE & $\textbf{88.3}$ & $\textbf{56.7}$ & $\textbf{94.0}$\\
		\cline{2-5}
		& ISL & $\textbf{88.9}$ & $\textbf{58.1}$ & $\textbf{94.5}$\\
		\hline\hline
		Architecture & Classifier/Feat. & \multicolumn{3}{c}{Linear Classifier / conv5}\\
		\hline
		\multirow{8}{*}{AlexNet} & Random & $67.3$ & $32.7$ & $79.2$\\
		\cline{2-5}
		& DeepCluster & $77.9$ & $41.9$ & $92.0$\\
		& RotNet & $\bm{84.1}$ & $57.4$ & $92.3$\\
		& Instance & $70.1$ & $39.4$ & $89.3$\\
		& AND & $77.6$ & $47.9$ & $\bm{93.7}$\\
		& ISL w/o HPE & $83.5$ & $\textbf{58.5}$ & $93.3$\\
		\cline{2-5}
		& PAD & $84.7$ & $58.6$ & $93.2$\\
		& ISL & $\textbf{85.8}$ & $\textbf{60.1}$ & $\textbf{93.9}$\\
		\hline
		\multirow{4}{*}{ResNet18} & Instance & $84.1$ & $48.9$ & $94.0$\\
		& AND & $88.9$ & $57.4$ & $94.3$\\
		& ISL w/o HPE & $\textbf{89.2}$ & $\textbf{61.1}$ & $\textbf{94.4}$\\
		\cline{2-5}
		& ISL & $\textbf{90.7}$ & $\textbf{63.5}$ & $\textbf{94.5}$\\
		\hline
		\multirow{4}{*}{ResNet50} & Instance & $85.0$ & $50.1$ & $94.4$\\
		& AND & $90.2$ & $58.5$ & $94.9$\\
		& ISL w/o HPE & $\textbf{91.0}$ & $\textbf{63.0}$ & $\textbf{94.9}$\\
		\cline{2-5}
		& ISL & $\textbf{91.5}$ & $\textbf{65.9}$ & $\textbf{95.2}$\\
		\hline
	\end{tabular}
	\vspace{-0.5cm}
\end{table}

\begin{table}[t]
	\footnotesize
	\caption{Comparison of top-1 accuracy (\%) on ImageNet with architectures of AlexNet, ResNet18 and ResNet50. The results of two classification models are reported: the weighted kNN with the FC features and the linear classifier using the Conv1-Conv5 features.}
	\label{imagenet}
	\centering
	\vspace{0.2cm}
	\renewcommand\arraystretch{1.1}
	\begin{tabular}{p{1.9cm}<{\centering}|p{0.55cm}<{\centering}p{0.55cm}<{\centering}p{0.55cm}<{\centering}p{0.55cm}<{\centering}p{0.55cm}<{\centering}|p{0.55cm}<{\centering}}
		\hline
		Classifier & \multicolumn{5}{c|}{Linear Classifier} & $k$NN\\
		\hline
		Feature & conv1 & conv2 & conv3 & conv4 & conv5 & FC\\
		\hline\hline
		\multicolumn{7}{c}{AlexNet}\\
		\hline
		Random & $11.6$ & $17.1$ & $16.9$ & $16.3$ & $14.1$ & $3.5$\\
		DeepCluster & $13.4$ & $32.3$ & $\bm{41.0}$ & $39.6$ & $38.2$ & $26.8$\\
		RotNet & $\bm{18.8}$ & $31.7$ & $38.7$ & $38.2$ & $36.5$ & $9.2$\\
		Instance & $16.8$ & $26.5$ & $31.8$ & $34.1$ & $35.6$ & $31.3$\\
		AND & $15.6$ & $27.0$ & $35.9$ & $39.7$ & $37.9$ & $31.3$\\
		PAD & - & - & - & - & $38.6$ & $35.1$\\
		LA & $18.7$ & $\bm{32.7}$ & $38.1$ & $42.3$ & $42.4$ & $38.1$\\
		ISL & $17.3$ & $29.0$ & $38.4$ & $\bm{43.3}$ & $\bm{43.5}$ & $\bm{38.9}$\\
		\hline\hline
		\multicolumn{7}{c}{ResNet18}\\
		\hline
		DeepCluster & $\bm{16.4}$ & $17.2$ & $28.7$ & $44.3$ & $49.1$ & $-$\\
		Instance & $16.0$ & $\bm{19.9}$ & $29.8$ & $39.0$ & $44.5$ & $41.0$\\
		LA & $9.1$ & $18.7$ & $\bm{34.8}$ & $48.4$ & $52.8$ & $45.0$\\
		ISL & $15.3$ & $19.1$ & $32.7$ & $\bm{49.1}$ & $\bm{54.0}$ & $\bm{46.1}$\\
		\hline\hline
		\multicolumn{7}{c}{ResNet50}\\
		\hline
		DeepCluster & $18.9$ & $27.3$ & $36.7$ & $52.4$ & $44.2$ & $-$\\
		LA & $10.2$ & $23.3$ & $39.3$ & $49.0$ & $60.2$ & $49.4$\\
		ISL &$\bm{17.3}$ & $24.2$ & $38.5$ & $52.5$ & $61.2$ & $\bm{50.2}$\\
		MoCo-v1$^{*}$ &$15.7$ & $22.9$ & $40.6$ & $50.8$ & $60.6$ & $37.7$\\
		MoCo-v2$^{*}$ &$14.9$ & $\bm{28.4}$ & $41.7$ & $\bm{52.9}$ &$67.5$& $38.5$ \\
		MoCo-v2+ISL$^{*}$ &$13.2$ & $27.1$ & $\bm{41.9}$ & $51.7$ &$\bm{68.6}$& $40.1$ \\
		\hline
	\end{tabular}
	\vspace{-0.5cm}
\end{table}

\subsection{Comparison with the State-of-the-art Methods}
In this section, we compare the proposed ISL with the state-of-the-art unsupervised representation learning methods including the clustering method DeepCluster\cite{caron2018deep}, the instance specificity analysis methods Instance \cite{wu2018unsupervised}, MoCo-v1 \cite{he2019momentum} and MoCo-v2 \cite{chen2020improved}, self-supervised methods RotNet \cite{gidaris2018unsupervised} and the neighborhood discovery methods AND \cite{huang2019unsupervised}, LA \cite{zhuang2019local}, PAD \cite{huang2020unsupervised}. Meanwhile, the baselines of random features are provided for reference. We demonstrate the top-1 accuracy on CIFAR-10, CIFAR-100, SVHN and ImageNet. 

For the experiments on CIFAR-10, CIFAR-100 and SVHN, we utilized AlexNet, ResNet18 and ResNet50 as the backbone network to evaluate the proposed ISL. We tested two classification models, the weighted kNN with the FC features and the linear classifier using the Conv5 features. Table \ref{small_dataset} demonstrates the results.  All the unsupervised learning methods outperform the random features by a sizable margin, which clearly shows the effectiveness. Except for PAD, other existing methods did not apply hard positive enhancement (HPE) strategy in unsupervised representation learning. As the hard positive enhancement (HPE) strategy also increases the accuracy of the learned representation, we also tested our ISL without HPE on the three datasets to evaluate the benefit brought only by the instance similarity learning, which is denoted as ISL w/o HPE in Table \ref{small_dataset}. Compared with existing unsupervised features, our ISL archives higher accuracy on all three datasets with the two classification models in most cases.

For experiments on ImageNet, AlexNet, ResNet18 and ResNet50 were applied as the backbone in our ISL. Despite of the kNN classification model with FC features was used for evaluation, the features from the Conv1 to Conv5 layers were also utilized to test our model as shown in Table \ref{imagenet}. The methods with the marker $*$ set the feature dimension as $2,048$. Due to the local aggregation metric that automatically pushes away dissimilar samples and pulls together similar instances, LA obtained the state-of-the-art performance among neighborhood discovery methods. However, LA ignored the mismatch between the Euclidean distance of sample pairs and the geodesic distance among instances that revealed the semantics. On the contrary, our ISL mines the feature manifold via GANs, and learns the instance similarity through the generated proxy to supervise the representation learning. Our method achieves the best performance among all existing neighborhood discovery methods when applying the high-level Conv4 and Conv5 features in the linear classifier and the FC features in kNN. MoCo-v1 \cite{he2019momentum} verified that building large and consistent dictionary on-the-fly via momentum contrast could facilitate effective largescale contrastive learning, and SimCLR \cite{chen2020simple} validated that an extra MLP projection head and more data augmentation benefited the contrastive learning. In order to further enhance the performance of our ISL, we integrated the proposed method with MoCo-v2 \cite{chen2020improved} that combined the techniques from both MoCo-v1 and SimCLR. The accuracy of Moco-v2 was obtained by rerunning the officially released code. Since our ISL employ the neighborhood discovery via the geodesic distance on the mined feature manifold, the feature discriminativeness is further strengthened by the informative pseudo supervision in contrastive learning. Figure \ref{fig:visualization} visualizes an example of positive sample mining via LA and our ISL in different rounds during training. LA treats instances with similar appearance including colors and shapes as positive samples and fails to distinguish the fine-grained difference among various classes. On the contrary, our method mines the feature manifold to assign similarity among instances and successfully finds the semantically similar samples even with different appearance.

\section{Conclusion}
In this paper, we have presented an instance similarity learning (ISL) method for unsupervised feature representation. The proposed ISL mines the feature manifold by GANs and learns the semantic similarity among instances by exploring the mined feature manifold, through which informative pseudo supervision is provided to learn discriminative features. Extensive experiments demonstrate the superiority of the proposed method compared with the state-of-the-art unsupervised features.

\section*{Acknowledgements}
This work was supported in part by the National Key Research and Development Program of China under Grant 2017YFA0700802, in part by the National Natural Science Foundation of China under Grant 61822603, Grant U1813218, and Grant U1713214, in part by a grant from the Beijing Academy of Artificial Intelligence (BAAI), and in part by a grant from the Institute for Guo Qiang, Tsinghua University.

{\small
\bibliographystyle{ieee_fullname}
\bibliography{egbib}
}

\end{document}